%% file: main.tex
\documentclass[conference]{IEEEtran}
\IEEEoverridecommandlockouts
\usepackage{cite}
\usepackage{amsmath,amssymb,amsfonts}
\usepackage{algorithmic}
\usepackage{graphicx}
\usepackage{textcomp}
\usepackage{xcolor}

\usepackage{multirow}
\usepackage[hang, tight]{subfigure}
\usepackage{wrapfig}
\usepackage{hyperref}
\usepackage{url}
\usepackage{comment}
\usepackage{color}
\usepackage{etoolbox}
\usepackage[ruled,vlined]{algorithm2e}
\usepackage{amsmath}
\usepackage[flushleft]{threeparttable}
\usepackage{booktabs}
\DeclareMathOperator*{\argmin}{argmin}

\def\BibTeX{{\rm B\kern-.05em{\sc i\kern-.025em b}\kern-.08em
    T\kern-.1667em\lower.7ex\hbox{E}\kern-.125emX}}
\begin{document}

\title{Joint Regularization on Activations and Weights for Efficient Neural Network Pruning}

\author{\IEEEauthorblockN{Qing Yang$^1$, Wei Wen$^1$, Zuoguan Wang$^2$ and Hai Li$^1$}
\IEEEauthorblockA{
\textit{$^1$Department of Electrical and Computer Engineering, Duke University, Durham, North Carolina, USA} \\
\textit{$^2$Black Sesame Technologies, Santa Clara, California, USA} \\
$^1$\{qing.yang21, wei.wen, hai.li\}@duke.edu, $^2$zuoguan.wang@bst.ai
}}

\maketitle

\begin{abstract}
With the rapid scaling up of deep neural networks (DNNs), extensive research studies on network model compression such as weight pruning have been performed for improving deployment efficiency. 
This work aims to advance the compression beyond the weights to neuron activations. 
We propose the joint regularization technique which simultaneously regulates the distribution of weights and activations. 
By distinguishing and leveraging the significance difference among neuron responses and connections during learning, the jointly pruned network, namely \textit{JPnet}, optimizes the sparsity of activations and weights for improving execution efficiency. 
The derived deep sparsification of JPnet reveals more optimization space for the existing DNN accelerators dedicated for sparse matrix operations. 
We thoroughly evaluate the effectiveness of joint regularization through various network models with different activation functions and on different datasets.
With $0.4\%$ degradation constraint on inference accuracy, a JPnet can save $72.3\% \sim 98.8\%$ of computation cost compared to the original dense models, with up to $5.2\times$ and $12.3\times$ reductions in activation and weight numbers, respectively. 
\end{abstract}


\input{intro.tex}
\input{related_works.tex}
\input{approach.tex}
\input{experiments.tex}
\input{discussion.tex}
\input{conclusion.tex}

\bibliographystyle{./IEEEtran}
\bibliography{./IEEEexample}

\end{document}

%% file: intro.tex
\section{Introduction}

Deep neural networks (DNNs) have demonstrated significant advantages in many real-world applications, such as image classification, object detection and speech recognition~\cite{he2016deep,redmon2016you,amodei2016deep}. 
On the one hand, DNNs are developed for improving performance in these applications, which leads to intensive demands in data storage, communication and processing. 
On the other hand, the ubiquitous intelligence promotes the deployment of DNNs in light-weight embedded systems that are only equipped with limited memory and computation resource.
To reduce the model size while ensuring the performance quality, 
weight pruning has been widely explored. 
The weights in small values are taken as redundant parameters and removed with little impact on the model accuracy \cite{han2015learning,park2016faster}.
Utilizing the zero-skipping technique \cite{han2016eie} while computing on sparse weight parameters can further save the computation energy. 
In addition, many specific DNN accelerators~\cite{albericio2016cnvlutin,reagen2016minerva} leverage the intrinsic sparse activation patterns of the rectified linear unit (ReLU) function. 
The approach, however, cannot be extended to those activation functions that lack intrinsic zeros, e.g., leaky ReLU. 

Although prior techniques achieved tremendous successes, merely focusing on the weights cannot lead to the best inference efficiency, the crucial metric in DNN deployment, for the following three reasons. 
First, existing weight pruning methods reduce the fully-connected (\textit{fc}) layer size dramatically, while there lacks a systematic method to achieve a comparable compression rate for convolution (\textit{conv}) layers. 
The \textit{conv} layers account for most of the computation cost and dominate the inference time in DNNs, whose performance is usually bounded by computation instead of memory accesses~\cite{jouppi2017datacenter,zhang2015optimizing}. 
The most essential challenge of speeding up DNNs is to minimize the computation cost, i.e., the intensive multiple-and-accumulate operations (MACs). 
Second, the weights and activations together determine the performance of a network. 
Our experiments show that the zero-activation percentage obtained by ReLU decreases after applying the weight pruning~\cite{han2016eie}.
Such a deterioration in activation sparsity could potentially eliminate the advantage of the aforementioned accelerator designs. 
Third, the activation in DNNs is not strictly limited to ReLU. 
Non-ReLU activation functions, such as leaky ReLU and sigmoid, do not have intrinsic zero-activation patterns.

In this work, we propose the joint regularization technique to minimize the computation cost of DNNs by pruning both weights and activations. 
Unlike the na\"ive solution by pruning weights and activations in sequence, joint regularization is an end-to-end solution that simultaneously learns the sparse connections and neuron responses. 
Dynamic activation masks and static weight masks are learned at the same time with the joint regularization. 
Through the learning on the different importance of neuron responses and connections, the jointly pruned network, namely \textit{JPnet}, can achieve a balance 
between activations and weights and therefore further improve execution efficiency. 
Moreover, the JPnet not only stretches the intrinsic activation sparsity of ReLU, but also targets as a generic solution for other activation functions, such as leaky ReLU. 
Our experiments on various network models with different activation functions and on different datasets show substantial reductions in MACs by the JPnet.
Compared to the original dense models,  JPnet can obtain up to $5.2\times$ activation compression rate, $12.3\times$ weight compression rate and eliminate $72.3\% \sim 98.8\%$ of MACs. 
Compared with merely adopting the weight pruning \cite{han2015learning}, JPnet can further reduce the computation cost by $1.3\times \sim 10.5\times$ in our experiments. 

%% file: related_works.tex
\section{Related Works}
\label{section:related}

\textbf{Weight pruning}
emerges as an effective compression technique in reducing the model size and computation cost of neural networks.  
A common approach of pruning the redundant weights in DNNs is to include an extra regularization term (e.g., the $\ell_1/\ell_2$ regularization) in the loss function~\cite{liu2015sparse,park2016faster} to constrain the weight distribution. 
Then the weights below a heuristic threshold will be removed. 
Afterwards, a certain number of finetuning epochs will be applied to recover the accuracy loss induced by the pruning.
In practice, the direct-pruning and finetuning stages can be carried out iteratively to gradually achieve the optimal tradeoff between the model compression rate and accuracy. 
To avoid erroneous removal of important weights in the na\"ive pruning and finetuning approach, a dynamic compression method was proposed to recover those pruned weights whose expected updates are larger than an empirical threshold in each training iteration \cite{guo2016dynamic}. 
Rather than using $\ell_1/\ell_2$ regularization to constrain the weight magnitude and distribution, $\ell_0$ regularization can be adopted as a stochastic binary mask on weights, which was proven to produce a higher sparsification level \cite{louizos2017learning}. 
These regularization-based weight pruning approaches demonstrated high effectiveness, especially for \textit{fc} layers~\cite{han2015learning}. 
However, these methods are heuristic and lack theoretical guarantee for the convergence and compression performance. 
Being theoretically proved, sparse variational dropout can be utilized on individual weights to realize all possible dropout rates~\cite{molchanov2017variational,neklyudov2017structured}. 
The objective of weight pruning can also be transformed as a non-convex optimization problem which is mathematically solvable using the alternating direction method of multipliers (ADMM)~\cite{zhang2018systematic}.
Again, finetuning is needed to recover the accuracy drop for the sparsified model obtained by ADMM.

Removing the redundant weights in structured forms, e.g., the filters and filter channels, has been widely investigated too. 
For example, structured pruning~\cite{wen2016learning} applies $group\ lasso$ regularization on weight groups in a variety of self-defined shapes and sizes. 
In~\cite{molchanov2016pruning}, the rankings of filters are indicated by the first-order Taylor series expansion of the loss function on feature maps.
The filters in low rankings are then removed. 
The filter ranking can also be represented by the root mean square or the sum of absolute values of the filter weights~\cite{mao2017exploring,yu2017scalpel}. 
A theoretical view on the importance of neurons/filters can be derived from the perspective of variational information bottleneck which minimizes the mutual information between layers~\cite{dai2018compressing}. 
The structured pruning methods do not require dedicated supports for random sparse matrix and thus are hardware-friendly for conventional computation platforms. 
However, these methods seldom achieve a weight compression rate as high as the element-wise pruning methods. 

\textbf{Activation sparsity}
has been utilized in DNN accelerator designs. 
The activation sparsity originating from ReLU accelerates DNN inference with reduced off-chip memory access and computation cost \cite{chen2016eyeriss,albericio2016cnvlutin,reagen2016minerva}. 
A simple technique to improve activation sparsity was explored by zeroing out small activations~\cite{albericio2016cnvlutin}. 
However, the increment of activation sparsity is very limited with a concern of accuracy loss.
Moreover, these works heavily relied on the zero activations of ReLU, which cannot be extended to other activation functions. 
Dropout-based methods were proposed to regulate activation sparsity and obtain sparse feature representation~\cite{ba2013adaptive,makhzani2015winner}. 
These techniques incur essential model modifications, e.g., adding a binary belief network overlaid on the original model. 
Some other studies were dedicated for feature map pruning in \textit{conv} layers by learning to recognize and remove redundant channels \cite{gao2018dynamic,ye2018rethinking}. 
Our proposed joint regularization is an orthogonal technique to feature map pruning by dealing with activation redundancy in a much finer granularity, i.e., element-wise. 

Generally, the model size compression is the main focus of weight pruning, while the regulation of activation sparsification focuses more on the intrinsic activation sparsity by ReLU or exploiting the virtue of sparse activation in the DNN training for better model generalization. 
\textit{In contrast, our proposed joint regularization aims to reduce the DNN computation cost and accelerate the inference by simultaneously optimizing weight pruning and activation sparsification.}

%% file: approach.tex
\section{Approach}

\subsection{Joint Regularization}

For an $L$-layer neural network represented by 
the weight set $\mathbb{S}_W=\{\mathbf{W}_i:i=1,\ldots, L\}$ where $\mathbf{W}_i$ denotes the weights of layer $i$,
given the dataset $\mathcal{\{X, Y\}}$, the $\mathbb{S}_W$ will be learned to minimize the loss function as follows: 
\begin{equation}
\label{eq_general_loss}
\begin{gathered}
Loss=\frac{1}{n}\sum_{k=1}^{n}\mathcal{L}(\mathbf{y}_k, \mathcal{D}(\mathbf{x}_k, \mathbb{S}_W)), \\
\mathbb{S}_W^\ast = \argmin_{\mathbb{S}_W}\ \{Loss\}, 
\end{gathered}
\end{equation}
where $\{\mathbf{x}_k, \mathbf{y}_k\}$ is the sampled input-output pair from $\mathcal{\{X, Y\}}$, and $n$ is the minibatch size. 
The nonlinear relationship of the network
is modeled as $\mathcal{D}(\cdot)$. 
The cross-entropy is usually adopted as the function $\mathcal{L}(\cdot)$ for multi-class problems. 
For the common weight pruning techniques, the optimization problem extends the loss function in Equation (\ref{eq_general_loss}) with a regularization term on $\mathbb{S}_W$ as 
\begin{equation}
Loss = \frac{1}{n}\sum_{k=1}^{n}\mathcal{L}(\mathbf{y}_k, \mathcal{D}(\mathbf{x}_k, \mathbb{S}_W))+\alpha\cdot \mathcal{R}^W(\mathbb{S}_W).
\end{equation}
$\mathcal{R}^W(\cdot)$ can be configured as the $\ell_0/\ell_1/\ell_2$ regularization on weights with a strength $\alpha$. 
The $\mathcal{R}^W(\cdot)$ focuses on the optimal weight compression, whereas $\mathbf{A}_{i-1}\cdot \mathbf{W}_i$ in layer $i$ is determined by both the activation $\mathbf{A}_{i-1}$ from the previous layer and the weights $\mathbf{W}_i$ of layer $i$.
We propose joint regularization on both weights and activations to minimize the computation cost and optimize the execution efficiency thereafter. 
Overall, the loss function will be represented as: 
\begin{equation}
\small
\label{eq_joint_reg}
Loss = \frac{1}{n}\sum_{k=1}^{n}\mathcal{L}(\mathbf{y}_k, \mathcal{D}(\mathbf{x}_k, \mathbb{S}_W))+\alpha\cdot \mathcal{R}^W(\mathbb{S}_W)+\beta\cdot \mathcal{R}^A(\mathbb{S}_A),
\end{equation}
where $\mathbb{S}_A=\{\mathbf{A}_i: i=1,\ldots, L-1\}$. 
$\mathbf{A}_L$ indeed is the model output, which is not included into the activation regularization $\mathcal{R}^A(\cdot)$. 
It's inappropriate to apply the $\ell_1/\ell_2$ regularization for activation, as the regularization may constrain the activation magnitude and hinder the feature learning process. 
Hence, we propose to adopt the $\ell_0$ regularization, which minimizes the number of the effective activations without disturbing their magnitudes. 
More specific, for each layer $i$, a binary mask $\mathbf{T}_i$ acting as an information filter is designed for the original activations $\mathbf{A}_{orig,i}$ 
such as
\begin{equation}
\mathbf{A}_{m,i} = \mathbf{A}_{orig,i}\odot \mathbf{T}_i,
\end{equation}
where $\odot$ is the element-wise multiply operation. 
The $\ell_0$ optimization problem on activations is therefore transformed as the derivation of optimal mask set $\mathbb{S}_T=\{\mathbf{T}_i: i=1,\ldots,L-1\}$. 

\subsection{Joint Pruning Procedure}

When implementing the joint regularization, we choose the $\ell_1$ regularization on weight distribution for its ease of gradient derivation while training. 
After combining with $\mathbb{S}_T$ for $\ell_0$ regularization on activations, the loss function in Equation (\ref{eq_joint_reg}) can be rewritten as: 
\begin{equation}
\label{eq_joint_final}
\begin{gathered}
Loss = \frac{1}{n}\sum_{k=1}^{n}\mathcal{L}(\mathbf{y}_k, \mathcal{D}(\mathbf{x}_k, \mathbb{S}_W, \mathbb{S}_T))+\alpha\cdot ||\mathbb{S}_W||_1, \\
\mathbb{S}_W^\ast,\ \mathbb{S}_T^\ast = \argmin_{\mathbb{S}_W,\ \mathbb{S}_T}\ \{Loss\}. 
\end{gathered}
\end{equation}
%
To overcome the non-differentiability of the $\ell_0$ regularization, we adopt a deterministic solution to obtain the proper $\mathbb{S}_T$. 
Noting that small weights in layers are learned to be pruned, activations with small magnitudes are taken as unimportant and masked out to further minimize inter-layer connections. 
Considering that neurons are activated in various patterns according to different input classes, we propose \textit{dynamic} masks for the activation pruning.
This is different from the \textit{static} masks in the weight pruning. 

\begin{figure}[t]
\centering
\includegraphics[width=1.0\columnwidth]{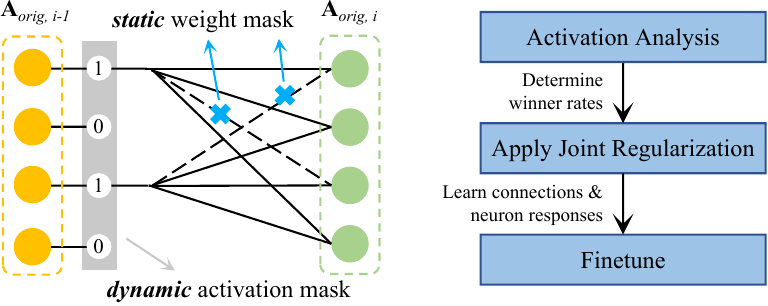}
\vspace{-10pt}
\caption{Working principle of joint pruning.}
\label{fig:how-to-prune}
\vspace{-10pt}
\end{figure}

\textcolor{black}{The selected activations by mask $\mathbf{T}_i$ are denoted as \textit{winners}. 
To derive the activation $a_{m,j} \in \mathbf{A}_{m,i}$ based on $a_{orig,j}\in \mathbf{A}_{orig,i}$, we have: 
\begin{equation}
\label{eq_mask_def}
a_{m,j} = 
\begin{cases}
a_{orig,j}, & \text{when}\ a_{orig,j}\ \text{is a winner}, \\
0, & \text{otherwise},
\end{cases}
\end{equation}
here the winners are dynamically determined at run-time according to the winner rate per layer.
The determination of winners through the activation mask is a relaxed \textit{partial sorting} problem to find top-$k$ arguments in an array.}
The winner rate of layer $i$ is defined as:
\begin{equation}
\label{eq_wr_def}
(\mathrm{winner\ rate})_i = \frac{|\mathbf{A}_{m,i}|}{|\mathbf{A}_{orig,i}|}, 
\end{equation}
where $|\mathbf{A}_{m,i}|$ and $|\mathbf{A}_{orig,i}|$ respectively denotes the number of winners selected by $\mathbf{T}_i$ and that of the original activations. 
Usually, different layer features a unique optimal winner rate. 
To get the appropriate winner rate per layer, the model with configurable activation masks is tested on a validation set sampled from the training set. 
Verified by our experiments, the size of the validation set can be similar to that of the test set.
The accuracy drop is taken as the indicator of the model sensitivity for the winner rate setting. 
The $(\mathrm{winner\ rate})_i$ is set empirically according to the tolerable accuracy loss. 
Examples of activation sensitivity analysis will be presented in Section \ref{section:discussion}. 
After deriving the winner rates, dynamic activation masks are configured as illustrated in Fig. \ref{fig:how-to-prune}. 

To understand the working scheme of the optimization problem defined by the Equation (\ref{eq_joint_final}), we focus on the operation of a single layer $i$: 
\begin{equation}
\mathbf{A}_{orig,i}=f_i(\mathbf{W}_i, \mathbf{A}_{m,i-1})=f_i(\mathbf{W}_i, \mathbf{A}_{orig,i-1}\odot \mathbf{T}_{i-1}),
\end{equation}
\textcolor{black}{where $f_i(\cdot)$ represents the function of layer $i$.} 
In the backpropagation phase, the partial derivative of the loss function on $\mathbf{A}_{orig,i}$ is propagated backwards: 
\begin{equation}
\frac{\partial Loss}{\partial \mathbf{A}_{orig,i-1}}=\frac{\partial Loss}{\partial \mathbf{A}_{orig,i}}\cdot \frac{\partial \mathbf{A}_{orig,i}}{\partial \mathbf{A}_{m,i-1}} \cdot 
\frac{\partial \mathbf{A}_{m,i-1}}{\partial \mathbf{A}_{orig,i-1}}.
\end{equation}
The term $\frac{\partial \mathbf{A}_{m,i-1}}{\partial \mathbf{A}_{orig,i-1}}$ is equal to $\mathbf{T}_{i-1}$, which means the backpropagation process is masked in the same way as the forward propagation. 
Thereafter, only the activated neurons will be updated. 
For the weight updating in a finetuning iteration, a small decay will be applied according to the setting of $\ell_1$ regularization on weights. 
Those weights smaller than the empirical threshold will be pruned out. 

As summarized in Fig. \ref{fig:how-to-prune}, the proposed end-to-end joint pruning approach consists of three steps. 
First, the significance of activations per layer is analyzed to determine the winner rates and define the pruning strength of each dynamic activation mask. 
Afterwards, the regularizations on both weights and activations are applied for the following finetuning stage. 
With the joint regulating force by $||\mathbb{S}_W||_1$ and activation masks $\mathbb{S}_T$ as defined in Equation~(\ref{eq_joint_final}), weights and activations are co-trained to obtain deep sparsification. 
\textcolor{black}{Through finetuning, the generated model is jointly optimized by dynamic sparse activation patterns and static compressed weights.} 

\subsection{Optimizer and Learning Rate}

We start the pruning process with several warm-up finetuning epochs to obtain the preliminary sparse patterns in both weights and activations with joint regularization. 
The same optimizer for training the original model is adopted. 
The learning rate is set as $0.1\times\sim 0.01\times$ smaller than the original setting. 
Our experiments show that Adadelta \cite{zeiler2012adadelta} usually brings the best performance in the following pruning process after the warm-up finetuning, especially for deep sparsified activations. 
Adadelta adapts the learning rate for each individual weight parameter. 
Smaller updates are performed on neurons associated with more frequently occurring activations, whereas larger updates will be applied for infrequent activated neurons.
Hence, Adadelta is beneficial for sparse weight updates, which commonly occur in our joint pruning method. 
During finetuning, only a small portion of weight parameters are updated because of the combination of sparse patterns in weights and activations. 
The learning rate for Adadelta is recommended to be reduced $0.1\times\sim0.01\times$ compared to the setting for training the original model. 

\subsection{Reconcile Dropout and Activation Pruning}

In DNN training, dropout layer is commonly added after large \textit{fc} layers to avoid over-fitting. 
The neuron activations are randomly chosen in the feedforward phase, and weights updates will be only applied on the neurons associated with the selected activations in the backpropagation phase. 
Thus, a random partition of weight parameters are updated in each training iteration. 
Similarly, the activation mask only selects a small portion of activated neurons and realize sparse weight updates. 
However, the over-fitting is still prone to happen because the selected neurons with winner activations are always kept and updated. 
Thus the random dropout layer is still needed. 
In \textit{fc} layers, the number of remaining activated neurons is reduced to $|\mathbf{A}_{m,i}|$ from $|\mathbf{A}_{orig,i}|$ as defined in Equation (\ref{eq_wr_def}). 
Similar to \cite{han2015learning} dealing with sparse \textit{fc} layer training, the dropout layer connected after the activation mask is suggested to be modified with the setting: 
\begin{equation}
(\mathrm{dropout\ rate})_i 
= C_d\cdot \sqrt{(\mathrm{winner\ rate})_i}\ ,
\end{equation}
where the constant $C_d$ is the dropout rate in the training process for original models. 
The activation winner rate is introduced to modify the dropout strength to balance over-fitting and under-fitting. 
The dropout layers will be directly removed in the inference stage. 

\subsection{Winner Prediction in Activation Pruning}
\label{section:winner-prediction} 

The dynamic activation pruning method increases the activation sparsity and maintains the model accuracy. 
As aforementioned, the determination of $\mathbf{A}_{m,i}$ through the activation mask is actually a relaxed partial sorting problem. 
According to the Master Theorem \cite{bentley1980general}, partial sorting can be fast solved in linear time $\mathcal{O}(N)$ on average through recursive algorithms, where $N$ is the number of elements to be partitioned. 
To further speed up, $\mathbf{A}_{m,i}$ can be predicted based on a down-sampled activation set. 
A threshold $\theta$ is derived by separating top-$\varepsilon k$ elements from the down-sampled activation set comprising $\varepsilon N$ elements with $\varepsilon$ as the down-sampling rate. 
\textcolor{black}{Then $\theta$ is applied to derive $a_{m,j}\in \mathbf{A}_{m,i}$ from $a_{orig,j}\in \mathbf{A}_{orig,i}$ as follows: 
\begin{equation}
\label{eq:threshold}
a_{m,j} = 
\begin{cases}
a_{orig,j}, & \text{when}\ \mathbf{abs}(a_{orig,j}) > \theta, \\ 
0, & \text{otherwise}.
\end{cases}
\end{equation}}

%% file: experiments.tex
\section{Experiments}
\label{section:experiments}

We evaluate he joint regularization on various models ranging from multi-layer perceptron (MLP) to deep neural networks (DNNs) on three datasets, MNIST, CIFAR-10 and ImageNet (Table \ref{tab:summary}). 
In ResNet-50 \cite{he2016deep} and wide ResNet-32 \cite{zagoruyko2016wide}, \textit{conv} layers account for more than 99\% computation cost and are our focus. 
All the evaluations are implemented in TensorFlow. 

\begin{table*}[]
\centering
\caption{Summary of JPnets}
\label{tab:summary}
\begin{tabular}{c|rrrrr||r}
Network     & MLP-3 & Lenet-4    & ConvNet-5    & AlexNet & ResNet-50  & ResNet-32  \\
\hline
Dataset             & MNIST & MNIST  & CIFAR-10 & ImageNet & ImageNet & CIFAR-10   \\
Activation Function & ReLU  & ReLU  & ReLU     & ReLU  & ReLU  & Leaky ReLU \\
\hline
Accuracy Baseline       & 98.41\% & 99.4\% & 86.0\%     & 57.22\%  & 75.6\% & 95.0\%    \\
Accuracy Joint Regularization   & 98.42\% & 99.0\% & 85.9\%  & 57.26\%  & 75.7\% & 94.6\%    \\
\hline
Activation Percentage  & 17.1\% & 5.5\% & 43.6\%   & 37.9\%  &  17.7\% & 30.8\%     \\
Weight Compression Rate   & 10$\times$   & 12.3$\times$ & 2.5$\times$   & 5.3$\times$  &  1.6$\times$ & 3.1$\times$ \\
MAC Percentage   & 3.65\% & 1.2\% & 27.7\%   & 25.2\%  & 19.1\%  & 11.5\% 
\end{tabular}
\end{table*}

\subsection{Overall Performance}

The compression results of JPnets on activations, weights and MACs are summarized in Table \ref{tab:summary}.
Our method can learn both sparse activations and sparse weights. 
Compared to original dense models, JPnets achieve $\textbf{1.4}\times\sim\textbf{5.2}\times$ activation compression rate and $\textbf{1.6}\times \sim \textbf{12.3}\times$ weight compression rate. 
As such, JPnets execute only $\textbf{1.2\%} \sim \textbf{27.7\%}$ of MACs required in dense models. 
The accuracy drop is kept less than 0.4\%, and for some cases, the JPnets achieve even better accuracy (e.g., MLP-3, AlexNet and ResNet-50). 

\begin{figure}[t]
\centering
\begin{minipage}{3.3in}
\centering
\subfigure[Comparison of non-zero activation percentage.]{
\includegraphics[width=0.9\columnwidth]{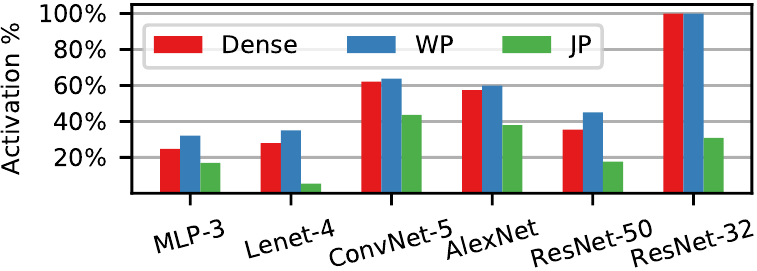}}
\subfigure[Comparison of MAC percentage.]{
\includegraphics[width=0.9\columnwidth]{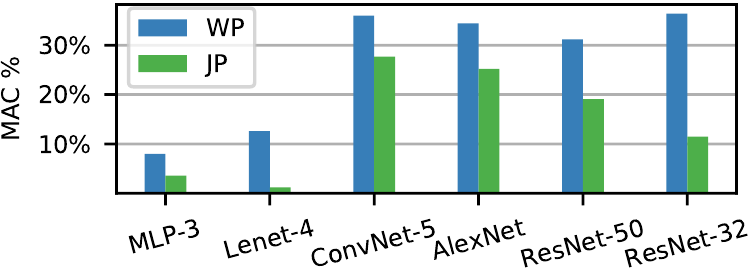}}
\caption{Comparison between WP and JP.}
\label{fig:compare-wp-ip}
\end{minipage}
\vspace{-10pt}
\end{figure}

The ReLU function in MLP-3, Lenet-4, ConvNet-5, AlexNet and ResNet-50 brings intrinsic zero activations.
However, our experiment results in Fig.~\ref{fig:compare-wp-ip}(a) show that the non-zero activation percentage in the weight-pruned (WP) model 
\textcolor{black}{tends to increase compared to the original dense models. 
This increment indeed undermines the benefit from weight pruning.}
Our proposed JP method can remedy the activation sparsity loss in WP models and remove $7.7\% \sim 22.5\%$ more activations even compared to the original dense models. 
We observe the largest activation removal in ResNet-32 which uses leaky ReLU as activation function. 
As leaky ReLU doesn't provide intrinsic zero activation, the WP model of ReNet-32 cannot benefit from activation sparsity. 
In contrast, the JPnet in this work can remove $69.2\%$ activations and reduce additional $25\%$ of MAC operations compared to the WP model. 
As shown in Fig. \ref{fig:compare-wp-ip}(b), JPnets decrease the MAC operations to $1.2\% \sim 27.7\%$. 
It is a $\textbf{1.3}\times \sim \textbf{10.5}\times$ improvement compared to WP models. 
More details on model configuration and analysis will be presented in the following subsections. 

\subsection{MNIST and CIFAR-10}

The MLP-3 on MNIST has two hidden layers with 300 and 100 neurons respectively.
The model configuration details are summarized in Table~\ref{tab:MLP}. 
The non-zero activation percentage (Acti \%) per layer indicates the pruning strength on activations before reaching next layer. 
The amount of MACs is calculated with batch size as 1. 
The same setting will be applied to the analysis for other models. 

The MLP-3 is successfully compressed $10\times$ and only $17.1\%$ of activations are kept. 
The total umber of MACs is reduced to merely $3.65\%$ ($27.4\times$) without compromising the model accuracy at all. 
A higher computation reduction rate is achieved by a Lenet-4 model which comprises two \textit{conv} layers and two \textit{fc} layers (Table \ref{tab:lenet4}). 
The JPnet for Lenet-4 reduces the computation cost $83.3\times$ with only $5.5\%$ activations and $8.1\%$ weights retained.

\begin{table}[]
\caption{MLP-3 on MNIST}
\label{tab:MLP}
\resizebox{0.9\columnwidth}{!}
{\begin{minipage}{\columnwidth}
\begin{tabular}{cc|cc|ccc}
Layer  & Shape & Weight \# & MAC \# & Acti \% & Weight \% & MAC \% \\
\hline
fc1   & 784$\times$300 & 235.2K     & 235.2K & 12\%   & 10\%       & 3.77\% \\
fc2   & 300$\times$100 & 30K        & 30K    & 24\%     & 10\%       & 2.62\% \\
fc3   & 100$\times$10  & 1K         & 1K     & 100\%     & 20\%       & 6.81\% \\
\hline
\multicolumn{2}{c|} {Total}   & 266.2K   & 266.2K & 17.1\%        & 10\%       & 3.65\%
\end{tabular}
\end{minipage}}
\end{table}

\begin{table}[]
\caption{Lenet-4 on MNIST}
\label{tab:lenet4}
\resizebox{0.9\columnwidth}{!}
{\begin{minipage}{\columnwidth}
\begin{tabular}{cc|cc|ccc}
Layer  & Shape & Weight \# & MAC \# & Acti \% & Weight \% & MAC \% \\
\hline
conv1   & 5$\times$5, 20 & 0.5K     & 0.4M & 6.6\%   & 60\%       & 11.5\%  \\
conv2   & 5$\times$5, 50 & 25K        & 4.9M    &  1.9\%    & 10\% & 0.7\% \\
\hline
fc1   & 2450$\times$500  & 1.23M         & 1.2M     &  12.2\%  & 8\%    & 0.2\%  \\
fc2 & 500$\times$10 & 5K & 5K & 100\% & 18\%  & 2.2\% \\
\hline
\multicolumn{2}{c|} {Total}   & 1.26M   & 6.5M &    5.5\%     & 8.1\%       & 1.2\%
\end{tabular}
\end{minipage}}
\end{table}

\begin{figure}[t]
\centering
\includegraphics[width=0.85\columnwidth]{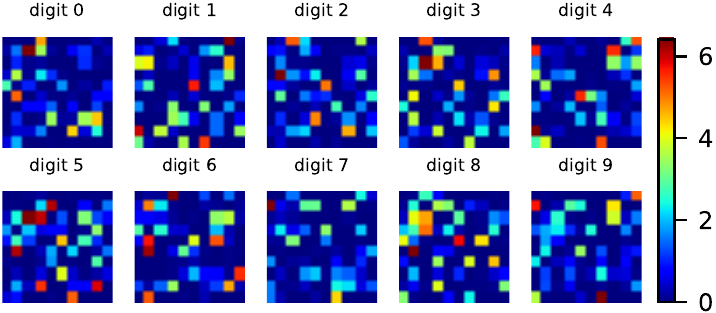}
\vspace{-6pt}
\caption{The activation distribution of \textit{fc2} in MLP-3 for all digits. The activation is obtained by averaging over all data examples per digit class.}
\label{fig:dist_all_digits}
\end{figure}

\begin{figure}[t]
\centering
\includegraphics[width=0.8\columnwidth]{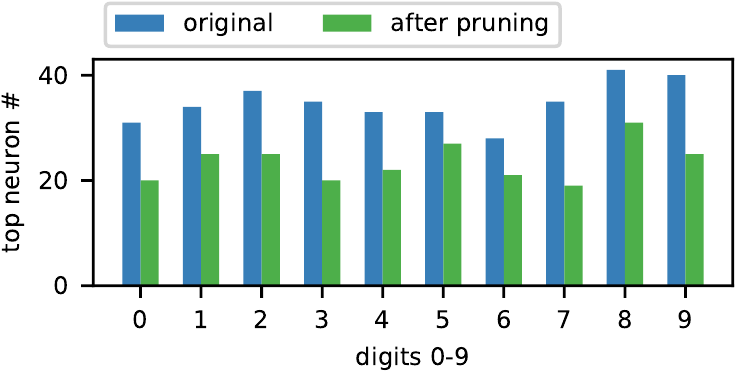}
\vspace{-6pt}
\caption{The number of activated top neurons for all digits.}
\label{fig:top-neuron-dist}
\vspace{-10pt}
\end{figure}

To analyze and understand the effectiveness of dynamic activation masks, we take the example of the activation patterns from layer \textit{fc2} in MLP-3. 
Before starting the joint pruning, the activation distribution for all MNIST digits is visualized in Fig. \ref{fig:dist_all_digits}, which clearly shows that digits $0-9$ incur different regions in \textit{fc2}. 
The observation implies that it is impossible to design a static activation mask and obtain a comparable sparsification effectiveness as the dynamic counterpart. 
We name the neuron featuring maximum activation for each input as \textit{top neuron}. 
Fig.~\ref{fig:top-neuron-dist} compares the number of activated top neurons for all digits by observing the training set before and after applying joint pruning. 
The results show that the JPnet needs fewer top neurons and generates a sparser feature representation. 

We also apply joint regularization to ConvNet-5 on CIFAR-10 dataset. 
The accuracy of the original dense model is $86.0\%$.
As detailed in Table \ref{tab:lenet}, JPnet for ConvNet-5 needs only $27.7\%$ of total MACs compared to the dense model by pruning $59.6\%$ of weights and $56.4\%$ of activations.
Only $0.1\%$ accuracy drop is resulted by JPnet. 
The \textit{conv} layers account for more than $80\%$ of total MACs and dominate the computation cost. 

\begin{table}[b]
\caption{ConvNet-5 on CIFAR-10}
\label{tab:lenet}
\resizebox{0.9\columnwidth}{!}
{\begin{minipage}{\columnwidth}
\begin{tabular}{cc|cc|ccc}
Layer & Shape              & Weight \# & MAC \# & Acti \% & Weight \% & MAC \% \\
\hline
conv1 & 5$\times$5, 64  & 4.8K       & 0.69M  & 50.6\%    & 70\%       & 70\%   \\
conv2 & 5$\times$5, 64 & 102.4K     & 3.68M  & 17.3\%   & 50\%       & 25.3\% \\
\hline
fc1   & 2304$\times$384    & 884.7K     & 884.7K & 9.9\%   & 40\%       & 6.92\% \\
fc2   & 384$\times$192     & 73.7K      & 73.7K  & 44.8\%   & 30\%       & 3.0\%    \\
fc3   & 192$\times$10      & 1.92K      & 1.92K  & 100\%   & 50\%       & 22.4\% \\
\hline
\multicolumn{2}{c|} {Total} & 1.07M      & 5.34M  & 43.6\%       & 40.4\%     & 27.7\%
\end{tabular}
\end{minipage}}
\end{table}

\subsection{ImageNet}
\label{sec:imagenet}

We use ImageNet ILSVRC-2012 dataset to evaluate the joint pruning method on large datasets. ImageNet consists of about 1.2M training images and 50K validating images. 
The AlexNet and ResNet-50 are adopted. 

\begin{figure}[t]
\centering
\begin{minipage}{3.3in}
\centering
\subfigure[The weight decomposition. Compression rate = $5.3\times$.]{
\includegraphics[width=0.45\columnwidth]
{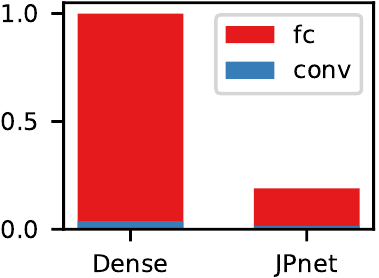}}
\hspace{0.1in}
\subfigure[The computation decomposition. Reduction rate = $4\times$.]{
\includegraphics[width=0.45\columnwidth]
{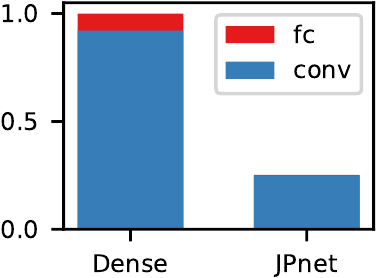}}
\caption{The decomposition of weight and computation cost of AlexNet. The overall model size and computation cost are normalized here.}
\label{fig:conv-vs-fc}
\end{minipage}
\vspace{-10pt}
\end{figure}

The AlexNet comprises 5 \textit{conv} layers and 3 \textit{fc} layers and achieves $57.22\%$ top-1 accuracy on the validation set. 
Similar to ConvNet-5, the computation bottleneck of AlexNet emerges in \textit{conv} layers, which accounts for more than $90\%$ of total MACs. 
As shown in Table \ref{tab:alexnet}, deeper layers present larger pruning strength on weights and activations due to the high-level feature abstraction of input images. 
For example, the MACs of \textit{conv5} can be reduced $18.2\times$, while only a $1.2\times$ reduction rate is realized in \textit{conv1}. 
In total, applying joint pruning removes $81.1\%$ weights and $62.1\%$ activations, inducing $4\times$ reduction in effective MACs. 
The weight and computation cost decomposition is shown in Fig. \ref{fig:conv-vs-fc}. 
The \textit{fc} layers contribute the most majority of model size and are generally pruned in larger strength than \textit{conv} layers to realize a significant model compression rate. 
Whereas, the computation cost reduction mainly comes from the optimization in \textit{conv} layers as depicted in Fig.~\ref{fig:conv-vs-fc}(b). 

\begin{table}[t]
\caption{AlexNet on ImageNet}
\label{tab:alexnet}
\resizebox{0.86\columnwidth}{!}
{\begin{minipage}{\columnwidth}
\begin{tabular}{cc|cc|ccc}
Layer & Shape                & Weight \# & MAC \# & Acti \% & Weight \% & MAC \% \\
\hline
conv1 & 11$\times$11, 96  & 34.85K     & 109.3M & 69.9\% & 85\%       & 85\%   \\
conv2 & 5$\times$5, 256  & 307.2K     & 240.8M & 28.6\%  & 40\%       & 27.9\% \\
conv3 & 3$\times$3, 384 & 884.7K     & 149.5M & 16.4\%  & 35\%       & 10\% \\
conv4 & 3$\times$3, 384 & 663.5K     & 112.1M & 13.7\%     & 40\%       & 6.6\%   \\
conv5 & 3$\times$3, 256 & 442.4K     & 74.8M  & 15\%     & 40\%       & 5.5\%   \\
\hline
fc1 & 9216$\times$4096  & 37.7M     & 37.7M  & 10\%     & 17.2\%       & 2.6\%  \\
fc2 & 4096$\times$4096  & 16.8M     & 16.8M  & 9.4\%     & 17.2\%       & 1.7\%  \\
fc3 & 4096$\times$1000  & 4M     & 4M  & 100\%     & 31\%   & 2.9\%  \\ 
\hline
\multicolumn{2}{c|} {Total}  & 60.9M      & 745.2M  & 37.9\%    & 18.9\%     & 25.2\%
\end{tabular}
\end{minipage}}
\end{table}

To reach higher accuracy, DNNs are getting deeper with tens to hundreds of \textit{conv} layers. 
We deploy the joint regularization on ResNet-50 and summarize the detailed results in Table \ref{tab:resnet-50}. 
Consisting of 1 \textit{conv} layer, 4 residual units and 1 \textit{fc} layer, the ResNet-50 model achieves a $75.6\%$ accuracy on ImageNet ILSVRC-2012 dataset. 
In each residual unit, several residual blocks equipped with \textcolor{black}{bottleneck layer} are stacked. 
The filter numbers in residual units increase rapidly, and the same for the weight amount as shown in the table. 
An average pooling layer is connected before the last \textit{fc} layer to reduce feature dimension. 
Overall, \textit{conv} layers contribute the most majority of weights and computation. 
The JPnet for ResNet-50 achieves a $75.7\%$ accuracy, which is $0.1\%$ higher than the original model. 
Only $19.1\%$ MACs are retained in JPnet with a $5.65\times$ activation reduction and a $1.6\times$ weight compression. 

\begin{table}[t]
\caption{ResNet-50 on ImageNet.}
\label{tab:resnet-50}
\resizebox{0.75\columnwidth}{!}
{\begin{minipage}{\columnwidth}
\begin{tabular}{cc|cc|ccc}
Layer & Shape   & Weight \# & MAC \# & Acti \% & Weight \% & MAC \% \\
\hline
conv1 & 7$\times$7, 64      &   9.4K    &  0.84G   &  39.9\%    &    91.5\%    &   91.5\% \\
\hline
unit2 & $\left \{ \begin{tabular}{l} 1$\times$1, 64 \\ 3$\times$3, 64 \\ 1$\times$1, 256 \end{tabular} \right \}$ $\times$ 3 & 0.21M  &  1.13G & 19.4\%  &   66.8\%    & 13.6\%  \\
unit3 & $ \left \{ \begin{tabular}{l} 1$\times$1, 128 \\ 3$\times$3, 128 \\ 1$\times$1, 512 \end{tabular} \right \} $ $\times$ 4 & 1.21M &  1.68G   &   19.6\%   &   68.2\%   &  14.3\%  \\
unit4 & $\left \{ \begin{tabular}{l} 1$\times$1, 256 \\ 3$\times$3, 256 \\ 1$\times$1, 1024 \end{tabular} \right \}$ $\times$ 6 & 7.08M  & 2.49G  &  12.7\%     &  59.1\%     & 8.4\%  \\
unit5 & $\left \{ \begin{tabular}{l} 1$\times$1, 512 \\ 3$\times$3, 512 \\ 1$\times$1, 2048 \end{tabular} \right \}$ $\times$ 3 &  14.94M & 1.49G   &  7.9\%    &   62.2\%    &  5.8\%  \\
\hline
\multicolumn{2}{c|} {Total}     &  25.5M    & 7.63G  &    17.7\%      &    61.6\%   &  19.1\%
\end{tabular}
\end{minipage}}
\end{table}

\subsection{Prune Activation without Intrinsic Zeros}
\label{section:resnet-32}

For the networks aforementioned, joint regularization stretches the sparsity level in the ReLU activation. 
In the following, we validate the idea on the activation function without intrinsic sparse patterns, e.g., leaky ReLU. 
Table \ref{tab:resnet-32} shows our results for ResNet-32. 
The model consists of 1 \textit{conv} layer, 3 stacked residual units and 1 \textit{fc} layer.
Each residual unit contains 5 consecutive residual blocks. 
Compared to \textit{conv} layers, the last \textit{fc} layer is negligible in terms of weight volume and computation cost. 
The original model has a $95.0\%$ accuracy on CIFAR-10 dataset with 7.34G MACs per image. 
As its activation function is leaky ReLU, zero activations rarely occur in the original and WP models. 
After applying joint pruning, the activation percentage can be dramatically reduced down to $30.8\%$. 
As shown in Table \ref{tab:resnet-32}, the JPnet keeps $32.3\%$ weight parameters, while only $11.5\%$ MACs are required in execution. 
The accuracy drop is merely $0.4\%$. 

\begin{table}[t]
\caption{ResNet-32 on CIFAR-10.}
\label{tab:resnet-32}
\resizebox{0.8\columnwidth}{!}
{\begin{minipage}{\columnwidth}
\begin{tabular}{cc|cc|ccc}
Layer & Shape   & Weight \# & MAC \# & Acti \% & Weight \% & MAC \% \\
\hline
conv1 & 3$\times$3, 16    & 0.43K      & 0.44M  &   50\%   &  40\%      &   40\% \\
\hline
unit2 & $\left \{ \begin{tabular}{c} 3$\times$3, 160 \\ 3$\times$3, 160 \end{tabular} \right \}$ $\times$ 5 & 2.1M       & 2.15G  &  29.1\%    &    40\%    &  11.6\%  \\
unit3 & $ \left \{ \begin{tabular}{c} 3$\times$3, 320 \\ 3$\times$3, 320 \end{tabular} \right \} $ $\times$ 5 & 8.76M      & 2.6G   & 31.8\%     &   40\%     & 12.7\%    \\
unit4 & $\left \{ \begin{tabular}{c} 3$\times$3, 640 \\ 3$\times$3, 640 \end{tabular} \right \}$ $\times$ 5 & 35.02M     & 2.6G   &  34.5\%    &   30\%      &  10.3\%  \\
\hline
\multicolumn{2}{c|} {Total}     & 45.87M     & 7.34G  &  30.8\%  & 32.3\%    & 11.5\%
\end{tabular}
\end{minipage}}
\vspace{-10pt}
\end{table}

\begin{figure}[t]
\centering
\begin{minipage}{3.3in}
\centering
\subfigure[Original.]{
\includegraphics[width=0.45\columnwidth]
{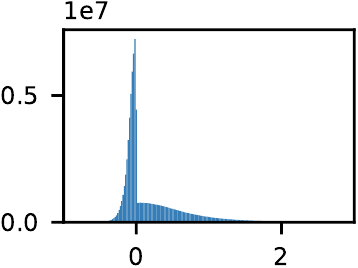}}
\hspace{0.1in}
\subfigure[After joint pruning.]{
\includegraphics[width=0.45\columnwidth]
{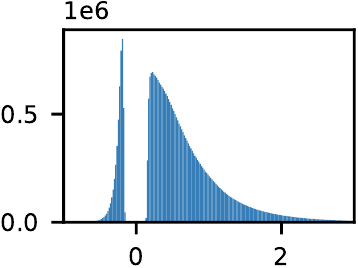}}
\caption{Activation distribution of ResNet-32.}
\label{fig:actis-distribution}
\end{minipage}
\end{figure}

Fig. \ref{fig:actis-distribution}(a) demonstrates the activation distribution of the first residual block in the original model by randomly selecting $500$ images from the training set. 
The distribution gathers near zero with long tails towards both positive and negative directions. 
For comparison, the activation distribution after joint pruning is shown in Fig. \ref{fig:actis-distribution}(b), in which activations near zero are pruned out. 
In addition, the kept activations are trained to be stronger with larger magnitude, which is consistent with the phenomenon that the non-zero activation percentage increases in WP models as illustrated in Fig. \ref{fig:compare-wp-ip}(a). 

%% file: discussion.tex
\section{Discussion}
\label{section:discussion}

\subsection{Comparison with Weight Pruning}

In Table \ref{tab:comparison}, we compare the joint pruning method with the state-of-the-art weight pruning methods, including $\ell_0$ regularization (L0) \cite{louizos2017learning}, variational information bottleneck (VIB) \cite{dai2018compressing}, variational dropout (VD) \cite{molchanov2017variational}, dynamic network surgery (DNS) \cite{guo2016dynamic} and the non-convex problem optimization method (ADMM) \cite{zhang2018systematic}. 
For the Lenet-4, our method achieves the best reduction rate on computation cost with similar inference error compared to others. 
While VIB and VD provide comparable reduction results on computation cost by merely focusing on weight pruning, the computational complexity during training hinders their application in DNNs for large datasets, e.g., ImageNet. 
Joint pruning can be easily applied for large models as shown in Section \ref{sec:imagenet}. 
Compared with DNS and ADMM, we can obtain the minimum prediction error with a comparable reduction rate on computation cost. 

\begin{table}[]
\caption{Comparison with the state-of-the-art weight pruning methods.}
\label{tab:comparison}
\begin{threeparttable}
\resizebox{1.0\columnwidth}{!}
{\begin{minipage}{\columnwidth}
\begin{tabular}{lllccc}
\toprule
Model        & Dataset          & Method & Weight \% & MAC Reduction & Error \\
\midrule
\multirow{5}{*}{Lenet-4} & \multirow{5}{*}{MNIST}    & L0     &       8.9\%    &    5.9$\times$           &     0.9\%  \\
  &             & VIB    &       0.8\%    & 71.4$\times$      &        1.0\% \\
  &             & VD     &      0.4\%     & 80.6$\times$           &    \textbf{0.8\%}    \\
  &             & Ours   &     8.1\%      &    \textbf{83.3$\times$}     &    1.0\%   \\
\midrule
\multirow{3}{*}{AlexNet$^\star$} & \multirow{3}{*}{ImageNet} & DNS    &      32.5\%     &     3.7$\times$          &    20\%   \\
 &       & ADMM   &     20.5\%      &   \textbf{3.8$\times$}            &     19.8\%  \\
 &       & Ours   &     38.7\%      &   3.7$\times$            &      \textbf{19.6\%} \\
\bottomrule
\end{tabular}
\begin{tablenotes}
\footnotesize
\item $^\star$ For AlexNet, we focus on \textit{conv} layers which are the computation bottleneck for inference. The top-5 prediction error is reported in the table. 
\end{tablenotes}
\end{minipage}}
\end{threeparttable}
\vspace{-10pt}
\end{table}

\subsection{Comparison with Static Activation Pruning}

The static activation pruning has been widely adopted in efficient DNN accelerator designs \cite{albericio2016cnvlutin,reagen2016minerva}. 
By selecting a proper static threshold $\theta$ in Equation (\ref{eq:threshold}), more activations can be pruned with little impact on model accuracy. 
For the activation pruning in joint pruning, the threshold is dynamic according to the winner rate and activation distribution layer-wise. 
The comparison between static and dynamic pruning is conducted on ResNet-32 for CIFAR-10 dataset. For the static pruning setup, the $\theta$ for leaky ReLU is assigned in the range of $[0.07, 0.14]$, which brings different activation sparsity patterns. 

\begin{figure}[t]
\centering
\includegraphics[width=0.75\columnwidth]{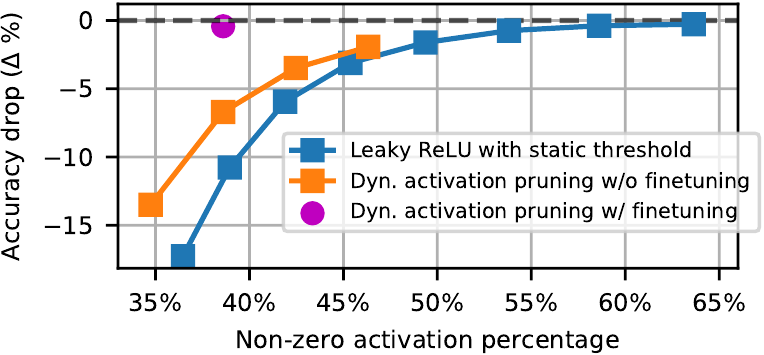}
\caption{Comparison to static activation pruning for ResNet-32.}
\label{fig:compare-to-static-relu}
\end{figure}

As the result of leaky ReLU with static threshold shown in Fig. \ref{fig:compare-to-static-relu}, the accuracy starts to drop rapidly when non-zero activation percentage is less than $58.6\%$ ($\theta=0.08$). 
Using dynamic activation masks, a better accuracy can be obtained under the same activation sparsity constraint. 
Finetuning the model using dynamic activation masks will dramatically recover the accuracy loss. As our experiment in Section \ref{section:resnet-32}, the JPnet for ResNet-32 can be finetuned to eliminate the $10.4\%$ accuracy drop caused by the static activation pruning. 

\subsection{Activation Analysis}

\begin{figure}[t]
\centering
\begin{minipage}{1.0\columnwidth}
\subfigure[AlexNet on ImageNet.]{
\includegraphics[width=0.45\columnwidth]{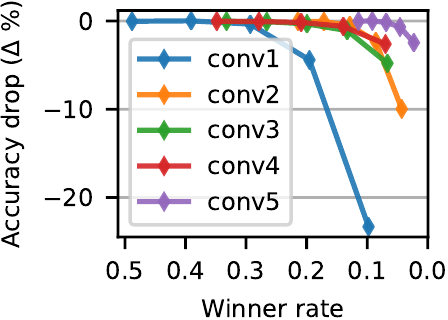}}
\hspace{0.1in}
\subfigure[ResNet-32 on CIFAR-10.]{
\includegraphics[width=0.45\columnwidth]{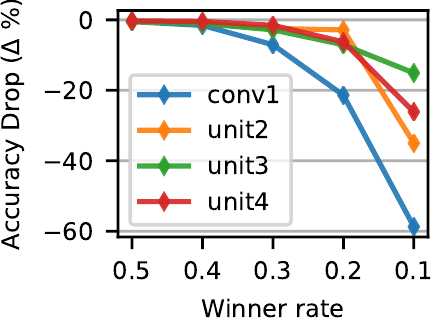}}
\caption{Activation pruning sensitivity.}
\label{fig:actis-pruning-sense}
\end{minipage}
\vspace{-10pt}
\end{figure}

In weight pruning, the applicable pruning strength varies by layers \cite{han2015learning,molchanov2016pruning}.
Similarly, the pruning sensitivity analysis is required to determine the proper activation pruning strength layer-wise, i.e., the activation winner rate per layer. 
Fig. \ref{fig:actis-pruning-sense}(a) shows the relation of JPnet accuracy drop and the selection of winner rates for AlexNet before pruning. 
As can be seen that the accuracy drops sharply as the activation winner rate of \textit{conv1} is less than $0.3$, while setting the winner rate of \textit{conv5} as $0.1$ doesn't affect accuracy. 
This implies that deeper \textit{conv} layers can support sparser activations. 
The unit-wise analysis results for ResNet-32 are shown in Fig. \ref{fig:actis-pruning-sense}(b), which denotes a similar trend of activation pruning sensitivity to AlexNet: 
\textit{conv1} is most susceptible to the activation pruning. 
The accuracy of ResNet-32 drops quickly with the decrements of winner rate, indicating a high sensitivity. 
Verified by thorough experiments in Section \ref{section:experiments}, the accuracy loss can be well recovered by finetuning with proper activation winner rates.

\subsection{Speedup from Dynamic Activation Pruning}

The speedup for \textit{fc} layers with dynamic activation pruning can be easily observed even without specific support for sparse matrix operations. 
After activation pruning, the weight matrix in \textit{fc} layers can be condensed by removing all connections related to the pruned activations, which speeds up the inference time with the compact weight matrix. 
Table \ref{tab:speedup} shows the experiment results implemented in TensorFlow compiled on Intel i7-7700HQ CPU for AlexNet's 3 \textit{fc} layers. 
The activation percentage listed here is the winner rate for the input activations. There is no accuracy loss after finetuning with these winner rate settings. 
Batch size is set as 1 in the test, which is the typical scenario in real-time applications on edge devices. 
The experiment obtains $1.95\times\sim3.65\times$ speedup. 
Time spent on activation pruning to get winner activations accounts for a very small portion of the time spent on the original densely connected layers. 

\begin{table}[t]
\caption{Speedup test for fc layers in AlexNet.}
\label{tab:speedup}
\resizebox{0.88\columnwidth}{!}
{\begin{minipage}{\columnwidth}
\begin{tabular}{c|c|c|c|c|c}
\multirow{2}{*}{Layer} & \multirow{2}{*}{Shape} & \multirow{2}{*}{Acti \%} & \multicolumn{2}{c|}{Time consumption per Layer}  & \multirow{2}{*}{Speedup} \\
\cline{4-5}
&                        &                          & Original & Acti Pruning + MACs &                    \\
\hline
fc1    & 9216$\times$4096       & 15\%  & 10.19 ms & 0.87 ms + 3.08 ms        & 2.58$\times$          \\
\hline
fc2    & 4096$\times$4096       & 10\%     & 4.54 ms  & 0.52 ms + 0.72 ms        & 3.65$\times$             \\
\hline
fc3    & 4096$\times$1000       & 9.4\%   & 1.52 ms  & 0.39 ms + 0.39 ms        & 1.95$\times$ 
\end{tabular}
\end{minipage}}
\end{table}

\subsection{Activation Threshold Prediction}

\begin{figure}[t]
\centering
\includegraphics[width=1.0\columnwidth]{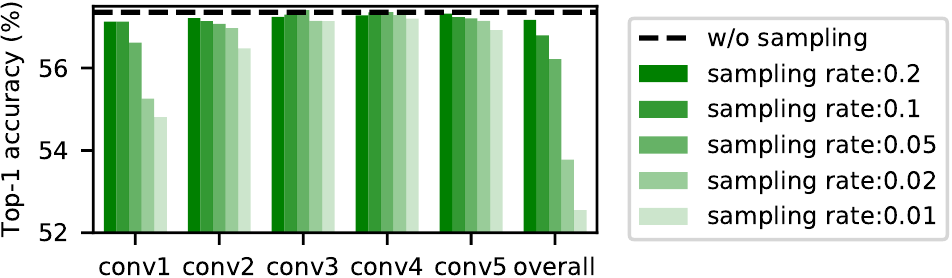}
\caption{The effects of threshold prediction.}
\vspace{-10pt}
\label{fig:wta-prediction}
\end{figure}

As discussed in Section \ref{section:winner-prediction}, the process to select activation winners can be accelerated by threshold prediction on down-sampled activation set. We apply different down-sampling rates on the JPnet for AlexNet. 
As can be seen in Fig. \ref{fig:wta-prediction}, layer \textit{conv1} is most vulnerable to threshold prediction.
From the overall results for AlexNet, it's practical to down-sample $10\%$ ($\varepsilon=0.1$) of activations for activation threshold prediction by keeping the accuracy drop less than $0.5\%$. 

%% file: conclusion.tex
\section{Conclusion}

To minimize the computation cost in DNNs, joint regularization integrating weight pruning and activation pruning is proposed in this paper. 
The experiment results on various models for MNIST, CIFAR-10 and ImageNet datasets have demonstrated considerable computation cost reduction. 
In total, a $1.4\times\sim 5.2\times$ activation compression rate and a $1.6\times\sim12.3\times$ weight compression rate are obtained. 
Only $1.2\% \sim 27.7\%$ of MACs are left with marginal effects on model accuracy, which outperforms the weight pruning by $1.3\times\sim 10.5\times$. 
The JPnets are targeted for the dedicated DNN accelerators with efficient sparse matrix storage and computation units on chip. 
The JPnets featuring compressed model size and reduced computation cost will meet the constraints from memory space and computing resource in embedded systems.